\documentclass[letterpaper, 10 pt, conference]{ieeeconf}  % Comment this line out if you need a4paper

\IEEEoverridecommandlockouts                              % This command is only needed if 
\usepackage{epsfig} % for postscript graphics files
\usepackage{times} % assumes new font selection scheme installed
\usepackage{amsmath} % assumes amsmath package installed
\usepackage{amssymb}  % assumes amsmath package installed
\usepackage{color}
\usepackage{xcolor}
\usepackage{preambles}
\usepackage{amsmath}
\usepackage{graphicx}
\usepackage{url} 
\usepackage[format=plain,labelsep=period]{caption}
\usepackage{wrapfig}
\usepackage[list=on,listformat=simple]{subcaption}
\usepackage{hyperref}
\usepackage{url}
\usepackage{caption}
\usepackage{float}
\floatstyle{plaintop}
\restylefloat{table}
\usepackage{subcaption}
\hypersetup{colorlinks = true, citecolor=black}
\title{\LARGE \bf
Safe Path Planning for Polynomial Shape Obstacles via Control Barrier Functions and
Logistic Regression}
\author{Chengyang Peng$^{1}$, Octavian Donca$^{1}$, and Ayonga Hereid$^{1}$% <-this % stops a space
\thanks{*This work was supported in part by the National Science Foundation under grant FRR-21441568. }% 
\thanks{$^{1}$Mechanical and Aerospace Engineering, Ohio State University, Columbus, OH, USA. {\tt\footnotesize (peng.947, donca.2, hereid.1)@osu.edu.}}%
}

\usepackage[capitalise]{cleveref}

\usepackage[noadjust]{cite}

\usepackage{algorithm2e,algorithmic}

\begin{document}

\maketitle
\thispagestyle{empty}
\pagestyle{empty}

\begin{abstract}
Safe path planning is critical for bipedal robots to operate in safety-critical environments. Common path planning algorithms, such as RRT or RRT*, typically use geometric or kinematic collision check algorithms to ensure collision-free paths toward the target position. However, such approaches may generate non-smooth paths that do not comply with the dynamics constraints of walking robots. It has been shown that the control barrier function (CBF) can be integrated with RRT/RRT* to synthesize dynamically feasible collision-free paths. Yet, existing work has been limited to simple circular or elliptical shape obstacles due to the challenging nature of constructing appropriate barrier functions to represent irregular-shaped obstacles. In this paper, we present a CBF-based RRT* algorithm for bipedal robots to generate a collision-free path through complex space with polynomial-shaped obstacles. In particular, we used logistic regression to construct polynomial barrier functions from a grid map of the environment to represent arbitrarily shaped obstacles. Moreover, we developed a multi-step CBF steering controller to ensure the efficiency of free space exploration. The proposed approach was first validated in simulation for a differential drive model, and then experimentally evaluated with a 3D humanoid robot, Digit, in a lab setting with randomly placed obstacles. 
% The results demonstrate 
% Robot motion planning is a fundamental and challenging problem because it often requires consideration of safety, efficiency, and compliance with robot dynamics. Rapidly Exploring Random Trees (RRT or RRT*) is one of the most common motion planning algorithms, which generates path by randomly sampling points in a map. However, this high-level planning and low level control synthesis may not be applicable to robots with complex dynamics. Control Barrier Function (CBF) based RRT/RRT* has enabled us to combine high-level planning and control synthesis. However, most research of CBF-RRT/RRT* focus on circular shape obstacles, which may not be an efficient representation of real-world environment. We present a CBF-RRT* based path planning algorithm that integrates with logistic regression. The proposed algorithm can generate a collision-free path to polygon or complex shape obstacles. Also, we develop a novel steering framework work to incorporate CBF constraints. The resulting path has demonstrated the safety and efficiency in multiple polygon shape obstacles in simulation. We also present the validation of the algorithm on Digit, a bipedal robot, in the real environment to verify the feasibility of the path.
\end{abstract}

\section{Introduction}
\label{sec:intro}

Mobile robots have shown encouraging promises in many real-world applications outside traditional well-structured factory settings thanks to the recent advancement of real-time path planning~\cite{gasparetto2015path}. 
Path planning has been extensively studied over the past decades~\cite{sleumer1999exact, xue2018solving}. A feasible path for a robot requires starting from an initial position to the goal position without colliding with any obstacle in the environment. Arguably the most prevailing approach in path planning is the sampling-based Rapidly Exploring Random Trees (RRT) algorithm, which expends the path by randomly sampling points in the configuration space~\cite{bruce2002real}. 
% However, the path generated by RRT is not optimal due to the probabilistic completeness. 
To improve the optimality of the resulting path, Karaman and Frazzoli~\cite{karaman2011sampling} proposed RRT*, which can reconnect the newly added node to the nearby nodes based on the minimum cost from the root node to the new node. 
% The limitation of the RRT* is that its control synthesis is low-level, which may not apply to some complex dynamics systems. 
Much progress has been made recently in combining low-level control synthesis and path planning, such as LQR-RRT*~\cite{perez2012lqr,goretkin2013optimal}, to ensure that the generated paths are consistent with the underlying dynamics constraints of the robot.
% However, limited by the collision check method, these algorithms are insufficient in path safety.
% Nowadays, human activities in many sectors can be supported or substituted by robots. Since operations at high speed or in a complex environment are required for modern mobile robots, such as low-altitude flight operations of drones, intra-regional transportation of wheeled robots, and efficient and safe production operations of industrial robots, robotics path planning has become a crucial research topic~\cite{montazeri2021unmanned}. As a class of mobile robots, biped robots also have the demands of path planning. Due to their kinematic capabilities, they can outperform wheeled robots in cluttered environments~\cite{hildebrandt2017real}. However, since the bipedal system is an unstable system~\cite{albert2001detection}. It is also the complex kinematics that makes the path planning of biped robots challenging. 

% As the basic but challenging problem in robotics, 

\begin{figure}
\vspace{2mm}
    \centering
    \includegraphics[trim={0cm 0cm 0cm 0cm},clip,width=\linewidth]{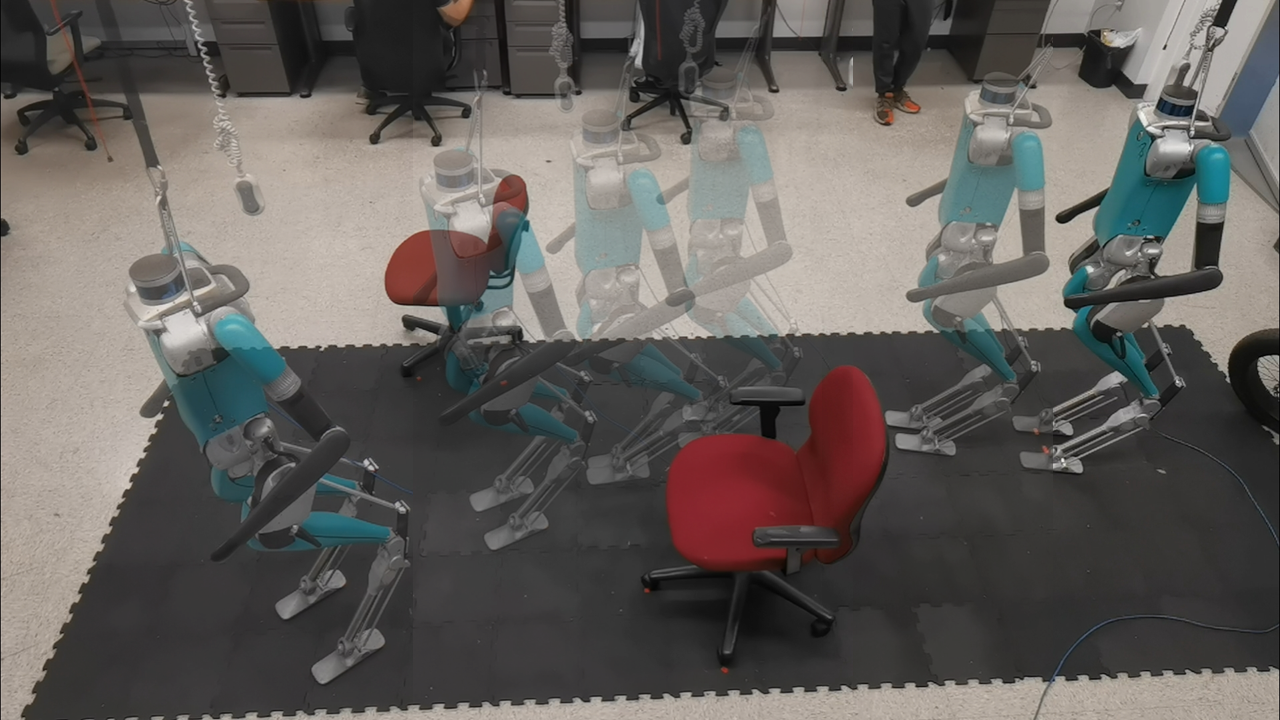}
    \caption{The snapshots of the bipedal robot, Digit, following the collision-free path generated by the proposed algorithm.}
    \label{fig:rd3}
\end{figure}

With the trending occasions of robots operating in the safety-critical environment (e.g., around people or in crowded spaces), the safety of robot motion becomes increasingly critical for the continuous deployment of these intelligent machines. 
Control Barrier Function, a popular tool in guaranteeing safety for nonlinear systems and constraints~\cite{xiao2022control}, has been shown effective in enforcing the safety-critical constraints on nonlinear systems such as autonomous vehicles and bipedal robot locomotion ~\cite{ames2019control,teng2021toward}. Recently, this method has also been used for designing safety-critical path planners. Yang et al. introduced a Quadratic Program (QP) that enforces Control Barrier Function (CBF) constraints to achieve obstacle avoidance~\cite{yang2019sampling}. Aniketh et al. proposed a framework to incorporate CBF constraints into the RRT path planning ~\cite{manjunath2021safe}. On these foundations, Ahmad et al. also combined RRT* algorithm with the CBF and equipped it with an adaptive sampling method to improve the efficiency~\cite{https://doi.org/10.48550/arxiv.2206.00795}. However, these obstacles studied by these algorithms only focused on circular and elliptical shapes because it would be easy to obtain their barrier functions. In many real-world scenarios, the circular barrier function is insufficient or wasteful to represent complex-shaped obstacle regions. 

In this work, we developed a modified CBF-RRT* algorithm with a new CBF-QP based multi-step steering controller for safe path planning in complex environments. The contributions of the proposed work are as follows. First, we proposed a new method that uses logistic regression to construct barrier functions that use polygon shapes to represent complex obstacles. Second, instead of calculating CBF-QP once when sampling a new node and moving one step, we would divide one step into four small steps and calculate them each, which can effectively keep the robot safe (avoiding collision). Finally, we applied our modified CBF-RRT* algorithm to bipedal robots to enable the robot to navigate safely in a room with complex obstacles and unreachable regions. We evaluated the proposed algorithm on a Digit robot in the lab setting and demonstrated safe navigation of bipedal walking robots.

The rest of the paper is organized as follows. \secref{sec:background} reviews the background of the control barrier function (CBF) and its integration with RRT/RRT* based planning algorithms. In \secref{sec:method}, we presents the core contribution of the paper, a CBF-RRT* planning algorithm with multi-step steering and polynomial-shaped barrier representation of complex obstacles. The simulation and experimental results with Digit robot are presented in \secref{sec:results}. Finally, \secref{sec:conclusion} briefly discusses the limitation of the proposed work and future research directions.

\SetKwFunction{CP}{ChooseParent}
\SetKwFunction{Rw}{Rewrite}

\section{Background}
\label{sec:background}
% In this section, we briefly review the mathematical background logistic regression, rapidly exploring random tree (RRT*) algorithm, and the control barrier function (CBF). 
In this section, we briefly review the mathematical basis of the control barrier function (CBF) and how it has been integrated with the RRT/RRT* based planning algorithms for safe navigation. 
% We will then briefly review the logistic regression algorithm, which will be used to construct polynomial-shaped barrier functions for complex obstacles from a grid map. 

\subsection{Control Barrier Function (CBF)}
We consider the robot dynamics can be written as the following affine nonlinear system:
\begin{align}
    \dot{x} = f(x)+g(x)u,
\end{align}
where $x \in \mathcal{X}$ is the system state with $\mathcal{X}\subseteq \mathbb{R}^n$ being the state space, and $u \in \mathcal{U}$ is the control input with $\mathcal{U}\subseteq \mathbb{R}^m$ being the control space. If there exists a continuous and differentiable function $h: \mathbb{R}^n \rightarrow \mathbb{R}$, the safety set $\mathcal{C}$ of the system can be defined as~\cite{nguyen2016exponential}:
\begin{equation}
\label{eq:safety-set}
\begin{aligned}
    \mathcal{C} &= \{x\in \mathbb{R}^n |h(x) \geq 0\},\\
    \partial\mathcal{C} &= \{x\in \mathbb{R}^n |h(x) = 0\},\\
    Int(\mathcal{C}) &= \{x\in \mathbb{R}^n |h(x) > 0\}.
\end{aligned}
\end{equation}
If $h(x)$ has relative degree $m > 1$, we can define a serious function $\psi_m (x): \mathbb{R}^n \rightarrow \mathbb{R}$ given as~\cite{xiao2019control}:
\begin{equation}
\begin{aligned}
    \psi_0 (x) &= h(x) \quad  &m = 0,\\
    \psi_m (x) &= \dot{\psi}_{m-1} (x)+\alpha_m (\psi_{m-1} (x)) \quad  &m \geq 1,
\end{aligned}
\end{equation}
where $\alpha_m(\cdot)$ is a class $\kappa$ function. The forward invariance safety condition can then be guaranteed if the following inequality constraints are satisfied for all $x\in\mathcal{C}$:
\begin{multline}
    \label{eq:cbf-cond}
    L_f^m h(x) + L_gL_f^{m-1}h(x)u+\frac{\partial^mh(x)}{\partial t^m}+O(h(x)) \\ 
    +\alpha_m(\psi_{m-1} (x)) \geq 0,
\end{multline}
where $O(h(x))$ denotes the remaining Lie derivatives along $f$ and partial derivatives with respect to $t$ with degree less than or equal to $m - 1$. Therefore, if $h(x)$ satisfied both \eqref{eq:safety-set} and \eqref{eq:cbf-cond}, it can be called a control barrier function. Since the control input $u$ is affine in \eqref{eq:cbf-cond}, one can formulate a quadratic programming (QP) controller subject to the CBF constraint in \eqref{eq:cbf-cond} to synthesize safe control actions~\cite{ames2019control,nguyen2016exponential,xiao2019control}.

% \begin{figure}
% \centering
% \vspace{2mm}
%      \begin{subfigure}[b]{1\columnwidth}
%          \centering
%         \includegraphics[trim={0.1cm 2.5cm 0.1cm 2.5cm},clip,width=0.9\columnwidth]{figures/Fig2new_a.png}
%          \caption{ChooseParent}
%          \label{fig:choosep}
%      \end{subfigure}
%      \begin{subfigure}[b]{1\columnwidth}
%          \centering
%         \includegraphics[trim={0cm 2.5cm 0cm 2.5cm},clip,width=0.9\columnwidth]{figures/Fig2new_b.png}
%          \caption{Rewrite}
%          \label{fig:rewrite}
%      \end{subfigure}
% \caption{\small{Demonstration of ChooseParent and Rewrite method}
% } 
% \label{fig:cp&re}
% \end{figure}

\subsection{CBF-RRT/RRT*}

% Different from RRT, RRT* can always converge to a relatively optimal solution if the running time is adequate, and two methods play important roles in path optimization ~\cite{noreen2016optimal}:

% \begin{itemize}
%     \item \CP: finds the near neighbor nodes around the new node and checks the cost of the new node through each neighbor node. Finally, it chooses the neighbor node that makes the cost minimum, as the parent node of the new node.
%     \item \Rw: reconnect each near neighbor nodes with new node, and checks the costs of these near nodes through the new node. Finally, it selects the optimal cost and rewrite the tree.
% \end{itemize}
% \figref{fig:cp&re} shows the demonstration of these two methods. Hence, in this work, we would create a new algorithm based on RRT*.
Built upon the standard RRT algorithm, Yang et al. developed the CBF-RRT path planning algorithm that uses CBF-QP~\cite{ames2019control} based safety-critical controllers to generate intermediate control actions to steer the robot away from the obstacles when approaching them~\cite{yang2019sampling}. The CBF controller replaces the collision check function in the traditional RRT algorithm while still ensuring safety. 
In \cite{manjunath2021safe}, Aniketh et al. improved the computational efficiency of CBF-RRT further by replacing CBF-QP with a random sampling of control actions that satisfy the barrier condition described in \eqref{eq:cbf-cond}. While it preserves the nature of random exploration by RRT, these approaches are often unable to generate (probabilistically) optimal paths. 
% And Aniketh et al. believe that this approach leads to an increase in computationally expensive. To reduce the computation amount of CBF-QP, Aniketh et al. uses a method of randomly 
% choosing controls and verifying that they satisfy CBF-QP instead of solving CBF-QP~\cite{manjunath2021safe}. 
% However, none of their trajectories are relatively optimal. 
To improve the optimality of the resulting path, Ahmad et al. combined CBF-QP with RRT* based on the work in \cite{yang2019sampling}, and improved the sampling efficiency through adaptive sampling based on the cross-entropy method (CEM)~\cite{https://doi.org/10.48550/arxiv.2206.00795}. It has been shown that RRT* yields a relatively optimal solution if given sufficient computation time. This is realized through two critical procedures described below~\cite{noreen2016optimal}:

\begin{itemize}
    \item \CP: finds the near neighbor nodes around the new node. If there is no obstacle collision between the new node and each near node, algorithm will compute the cost of the new node through each near node. Finally, it chooses the neighbor node that makes the cost minimum, as the parent node of the new node.
    \item \Rw: reconnects each near neighbor node with the new node and check their collisions. Calculates the costs of these near nodes through the new node. Finally, it selects the optimal cost and rewrites the tree.
\end{itemize}
% \figref{fig:cp&re} shows the demonstration of these two methods. 
In~\cite{https://doi.org/10.48550/arxiv.2206.00795}, the authors replaced the collision check function in the above two procedures with a CBF-QP based steering function, which inevitably increased the computational overhead of the CBF-RRT*. It is also important to note that the three aforementioned studies expand the tree by randomly sampling a note on the tree to extend toward the target position. This practice is inefficient in expanding the tree outward into feasible areas, thereby increasing the total number of iterations, as well as the computation overhead, required for the algorithm.
Moreover, determining a proper set of barrier functions to describe obstacles and unreachable areas remains challenging when using CBF for sampling-based path planning. The existing work only considers simple shapes, such as circles or ellipses. 

% extends from randomly selected nodes in the tree. This method is easy to gather the trajectories in one place, but it is difficult to extend outwards. With the increase in iterations, the efficiency of the algorithm decreases as well.

\SetKwFunction{CP}{ChooseParent}
\SetKwFunction{Rw}{Rewrite}
\SetKwFunction{Sp}{Sampling}
\SetKwFunction{Near}{Nearest}
\SetKwFunction{At}{Atan2}
\SetKwFunction{St}{Steer}
\SetKwFunction{AD}{AddNode}
\SetKwFunction{FindN}{NearIndex}
\SetKwFunction{NG}{NearGoal}
\SetKwFunction{GP}{GetFinalPath}
\SetKwFunction{CC}{NoCollision}
\SetKwInput{Input}{Input}
\SetKwInput{Init}{Initialization}
\RestyleAlgo{ruled}

\section{Safe Navigation via Multi-Step CBF-QP Steering with RRT*}
\label{sec:method}
% In this section, we propose the designing of the path planner that incorporates logistic regression and control barrier function into RRT* framework. Firstly, we introduce how the barrier function can be generated. Then, the CBF-QP is obtained by using the barrier function we get, and is incorporated with RRT*. The detail of the steering method and structure of the algorithm is shown in the last.

In this section, we present a safe path planning algorithm for bipedal robots that integrates the control barrier function with RRT* to provide guaranteed obstacle avoidance without explicit collision checking. Moreover, we propose to construct polynomial barrier functions to represent complex obstacles or unreachable regions using logistic regression on the planar grid map of the environment. 
Finally, we develop a multi-step CBF steering algorithm to address the infeasibility issues that the state may end up in the unsafe set.

\subsection{Bipedal Path Planning with Simplified Model}

With the purpose of finding an obstacle-free path, we consider the bipedal robot as a simple mass, assuming that there exists a stable low-level locomotion control that can follow waypoint or velocity commands. While a biped robot can walk in all directions, we only consider forward walking and turning in this paper. This is due to the difficulty of accurately controlling lateral walking speeds. The robot swings left and right while walking sideways.  
% Also, due to the hard control and unstable locomotion, we don't consider sidewalk of the robot. 
Hence, for the path planning purpose, we regard the bipedal robot as a differential drive type model, with states given by:
\begin{align}
    \mathbf{x} &= [x_1, x_2, \theta]^T,\\
    \dot{\mathbf{x}} &= [v\cos{\theta}, v\sin{\theta}, \omega]^T,
\end{align}
where $(x_1,x_2,\theta)$ corresponds to robot’s position and heading direction, and $(v,\omega)$ represents the robot’s forward and angular velocity. If we assume $v$ is a constant, the system of the robot can be simplified as
\begin{align}
\label{eq:system_dynamics}
    \dot{\mathbf{x}} = \begin{bmatrix} v_1 \\ v_2\\ \omega \end{bmatrix} = \begin{bmatrix} v\cos{\theta} \\ v\sin{\theta}\\ 0 \end{bmatrix} + \begin{bmatrix} 0 \\ 0\\ 1 \end{bmatrix}\omega = f(x) + g(x)u,
\end{align}
with $u = \omega$. To avoid a collision with the obstacle, one needs to synthesize angular velocity commands that safely steer the robot away from the obstacle. 
% where
% \begin{align}
%     f(x) = \begin{bmatrix} v\cos{\theta} \\ v\sin{\theta}\\ 0 \end{bmatrix}, \text{and } g(x) = \begin{bmatrix} 0 \\ 0\\ 1 \end{bmatrix}
% \end{align}

\subsection{Construct Polynomial Barrier Functions via Logistic Regression from 2D Obstacle Map}

\begin{figure}
\centering
\vspace{2mm}
\centering
\includegraphics[trim={0cm 0cm 0cm 0cm},clip,width=0.9\columnwidth]{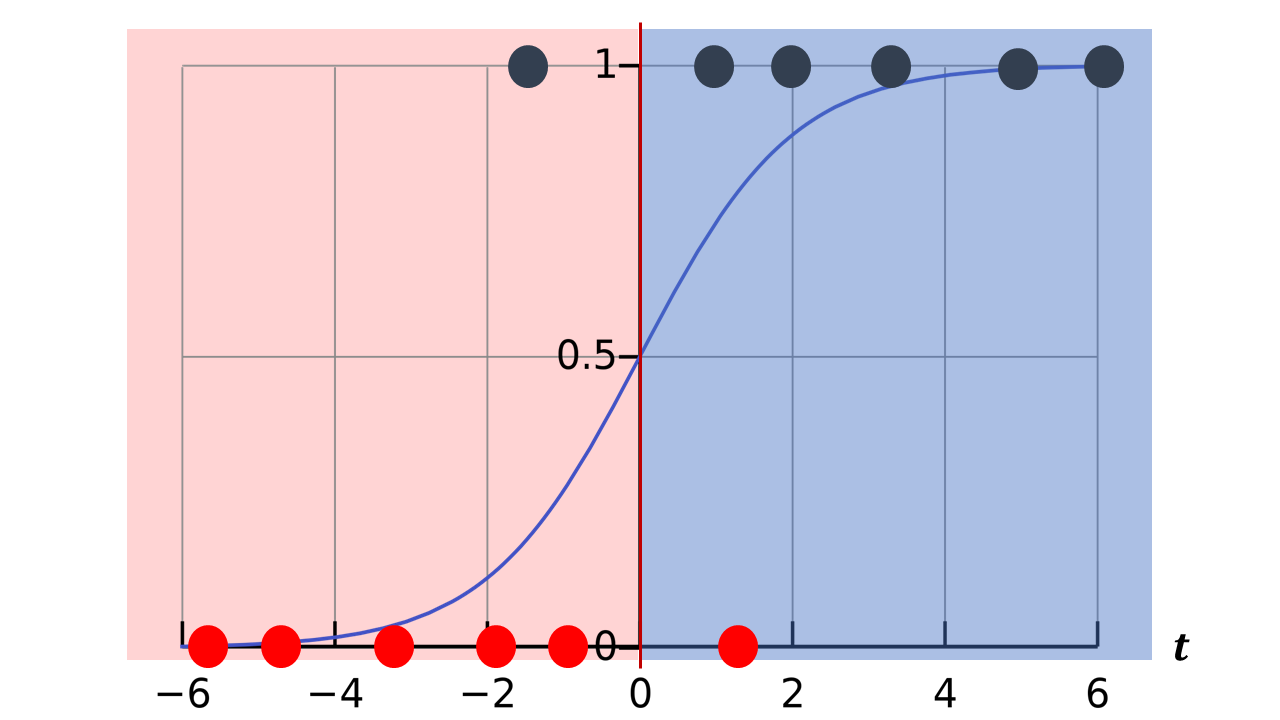}
\caption{\small{Two sets of data. Label 0 data is in red, and Label 1 is in gray. The sigmoid function is helpful in classifying two sets of data into two regions. In the blue region($t>0$), the label of data is more likely to be 1. In the red region($t<0$), the label of data is more likely to be 0. And $t=0$ becomes a boundary that separates these two regions.}
} 
\label{fig:Logistic regression}
\end{figure}

% \subsection{Logistic Regression}
To avoid collisions using CBF, we need to define a set of barrier functions $h(\mathbf{x})$ to represent the boundary of obstacles in the environment.
In this paper, we propose to use logistic regression to naturally construct polynomial-shaped barrier functions from a grid map. Compared to circular or ellipse shapes, polynomial shapes can more efficiently represent arbitrarily-shaped obstacles. Since the free space (i.e., safe set) is defined as the outside of the closed shapes, polygons can be used to represent obstacles when using CBF. 

Logistic regression is a standard probabilistic statistical classification model, which is used in classifying data with different labels~\cite{feng2014robust}. Different from linear regression, the outcome of logistic regression on one data sample is the probability of belonging to label 1 or label 0. The classification model (Sigmoid function) is given by:
\begin{align}
\label{eq:sigmoid func}
    \mathbb{P}(y=1| t)=\frac{1}{1+e^-t},
\end{align}
where $\mathbb{P}(y=1| t)$ represents the probability of the label of the feature $t$ is 1. Given \eqref{eq:sigmoid func}, $\mathbb{P}(y|t)\in [0.5,1)$ if $t>0$, meaning the label is more likely be 1; and $\mathbb{P}(y|t)\in[0,0.5)$ if $t<0$, meaning that the label is less likely be 1, or more likely be 0. Assuming that we have a set of data whose label could be either 1 or 0, as shown in \figref{fig:Logistic regression}, we can use $t=0$ as a classifier or decision boundary that can classify this set of data into two clusters. 
In particular, one can express $t$ as an affine function of a set of variables $\mathbf{z}= (z_0, z_1, z_2, ...,z_{j-1}, z_j)\in\R^j$, given as:
\begin{equation}
\label{eq:barrier}
    t = \boldsymbol{\beta} \mathbf{z}^T,
\end{equation}
where $\boldsymbol{\beta} = (\beta_0, \beta_1, \beta_2, ...,\beta_{j-1}, \beta_j)\in\R^j$ is a vector of the unknown coefficient of the function.
% \begin{align}
%     \boldsymbol{\beta} &= [\beta_0, \beta_1, \beta_2, ...,\beta_{j-1}, \beta_j]\\
%     \mathbf{z} &= [z_0, z_1, z_2, ...,z_{j-1}, z_j]
% \end{align}
By giving $N$ sets of data $\mathbf{z}$ and their label $\mathbf{y}$, the parameters $\boldsymbol{\beta}$ can be determined by minimizing the binary cross entropy cost:
\begin{align}
\label{eq:gradient_decend}
    \boldsymbol{\beta} = \argmin\sum_{n=1}^{N}(\ln(1+e^{t_i})-y_it_i)\quad \text{for } t_i = \boldsymbol{\beta}\mathbf{z_i}^T.
\end{align}
In this paper, we used Broyden–Fletcher–Goldfarb–Shanno (BFGS) algorithm, a method for solving unconstrained nonlinear optimization problems, to find $\boldsymbol{\beta}$.

\begin{figure}
\centering
% \vspace{2mm}
     \begin{subfigure}[b]{1\columnwidth}
         \centering
        \includegraphics[trim={0.1cm 2.5cm 0.1cm 2.5cm},clip,width=\columnwidth]{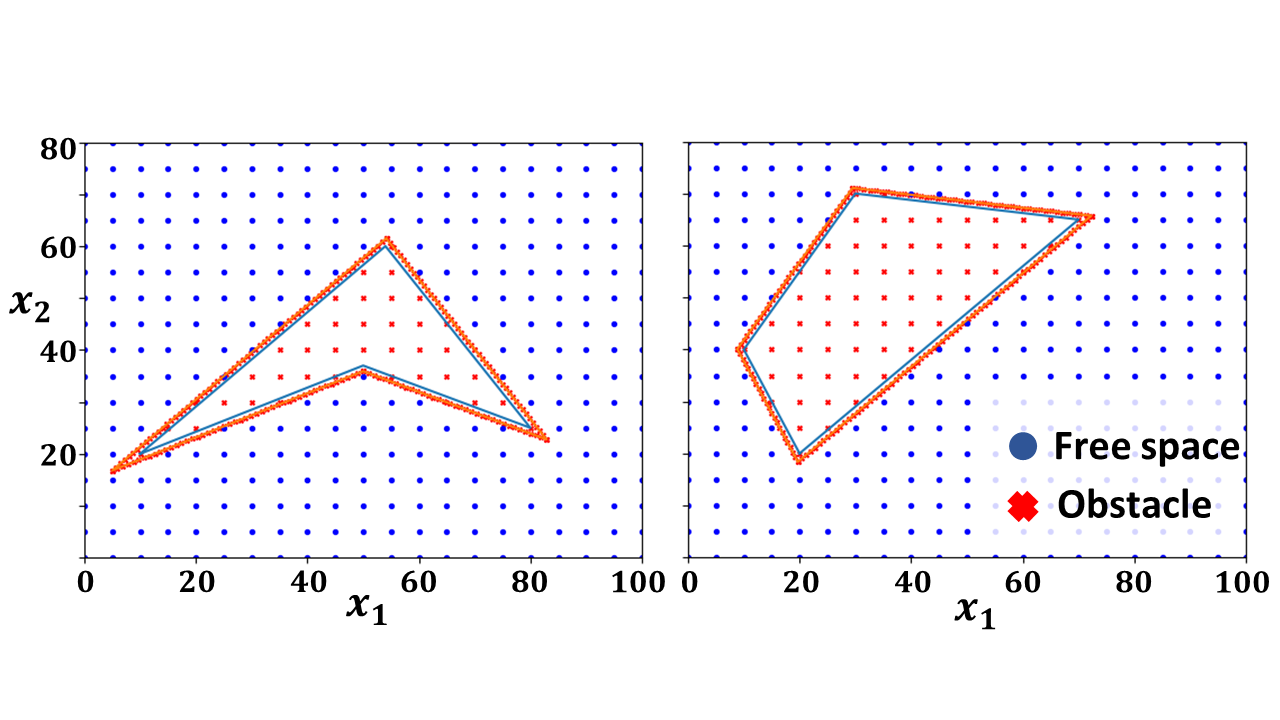}
         \caption{}
         \label{fig:Samp}
     \end{subfigure}
     \begin{subfigure}[b]{1\columnwidth}
         \centering
        \includegraphics[trim={0cm 2.5cm 0cm 2.5cm},clip,width=\columnwidth]{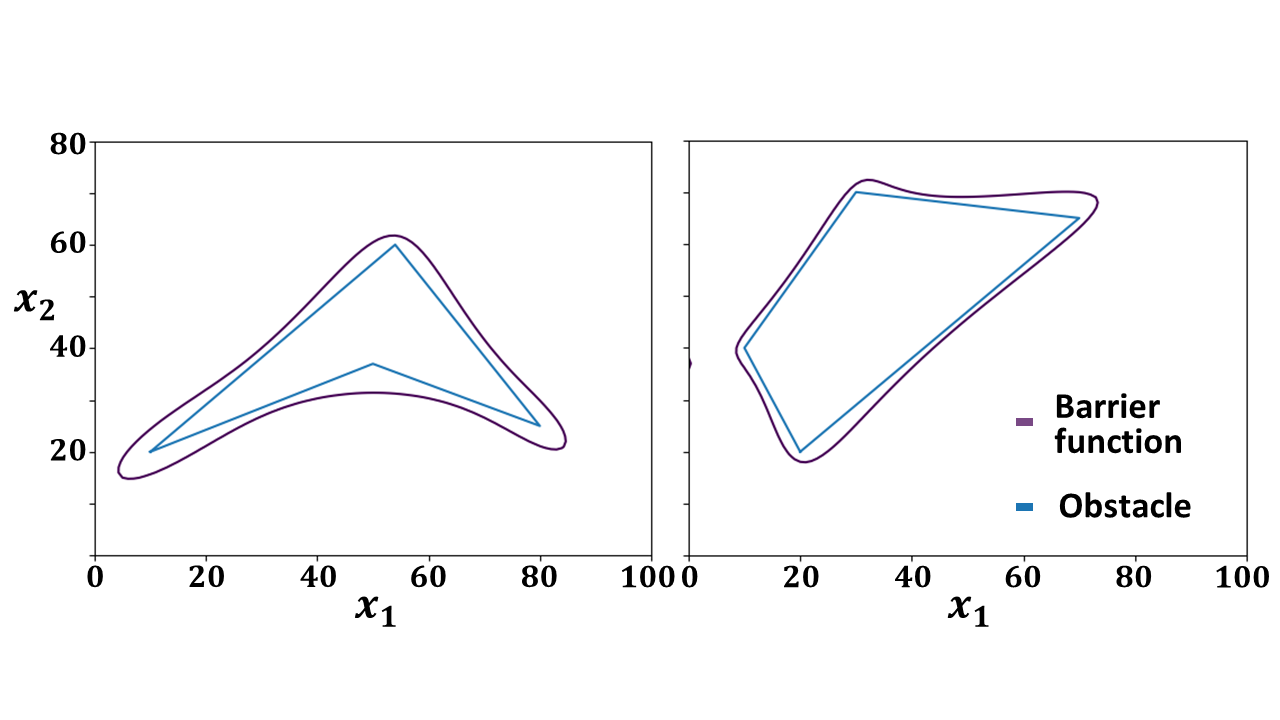}
         \caption{}
         \label{fig:barrier}
     \end{subfigure}
\caption{\small{Illustration examples of constructing barrier functions from 2D obstacle map. We first sampled equidistant $n$ points on the map and labeled each point either free or obstacle. (a). The sampled points in a map. Red crosses mean Label 0: when the point is in an obstacle; and Blue dots mean label 1: which is free space. (b). The generated barrier functions closely represent concave and convex obstacle regions.}
} 
\label{fig:barrier exp}
\end{figure}

The equation $t = \boldsymbol{\beta}\mathbf{z}^T = 0$ is the decision boundary of two sets. For our purpose, we can consider areas that have $t>0$ as free space (i.e., safe set), and $t<0$ as obstacle space (i.e., unsafe set). This is consistent with the CBF definition in \eqref{eq:safety-set}. To construct barrier functions from a 2D obstacle map, the first step is to determine the set of variables $\mathbf{z}$ from the robot's state variables $\mathbf{x}$. In this paper, we empirically select a set of polynomial functions of the robot's position with the maximum power of 4, given by
\begin{align}
\label{eq:states}
\mathbf{z} = [1, x_1^1x_2^0, x_1^0x_2^1, ...,x_1^1x_2^3, x_1^0x_2^4]
\end{align}
where $(x_1,x_2)$ is the position in the map. The coefficient vector is also defined accordingly, as $\boldsymbol{\beta} = [\beta_0, \beta_1, \beta_2, ...,\beta_{14}, \beta_{15}]$. A barrier function is then given by $h(\mathbf{x}) := \boldsymbol{\beta}\mathbf{z}^T$, which encloses an occupied area in the map.
% Therefore, we collect data for each point in a map and distinguish whether the points are in the obstacles or not, and we can generate suitable barrier functions to cover the obstacles. 

% For a map with a single polygon shape obstacle,  equidistant sampling the $n$ points in the map and distinguishing whether each of them is in obstacle or not. Label the points in the free space as 1 and obstacle as 0, and save this data into array $\mathbf{Y}$, see \figref{fig:Samp}. 
% Due to the 2D map and obstacle, should be closed. Therefore, $h(\mathbf{x})$ can be designed as a polynomial function with 4 as the maximum power:

 To solve the parameters $\boldsymbol{\beta}$, we need to get arrays by putting the position of each point into (\ref{eq:states}), and save these arrays into matrix $\mathbf{Z}$. Derive the cost function for $h(\mathbf{x})$, and using array $\mathbf{Y}$ which stores the label (occupied or free) of corresponding position and matrix $\mathbf{Z}$ to minimized the cost function given in (\ref{eq:gradient_decend}). Finally, using BFGS algorithm, we solve the parameters $\boldsymbol{\beta}$ to construct barrier functions. \figref{fig:barrier exp} shows two examples of constructing barrier boundaries of convex and concave-shaped obstacles.
 % . The algorithm also works for concave shape obstacle.

\begin{figure*}
\centering
\vspace{2mm}
     \begin{subfigure}[b]{0.32\linewidth}
         \centering
        \includegraphics[trim={0cm 0cm 0cm 0cm},clip,width=\linewidth]{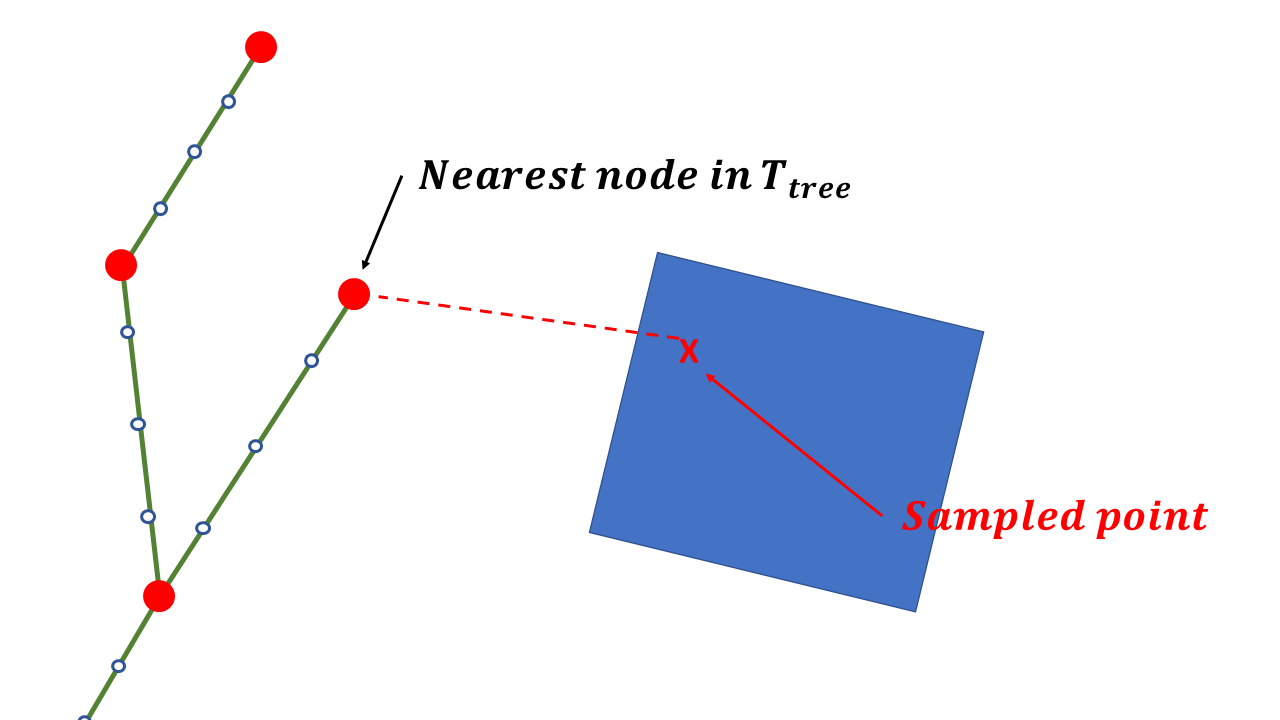}
         \caption{}
         \label{fig:step0}
     \end{subfigure}
     \begin{subfigure}[b]{0.32\linewidth}
         \centering
        \includegraphics[trim={0cm 0cm 0cm 0cm},clip,width=\linewidth]{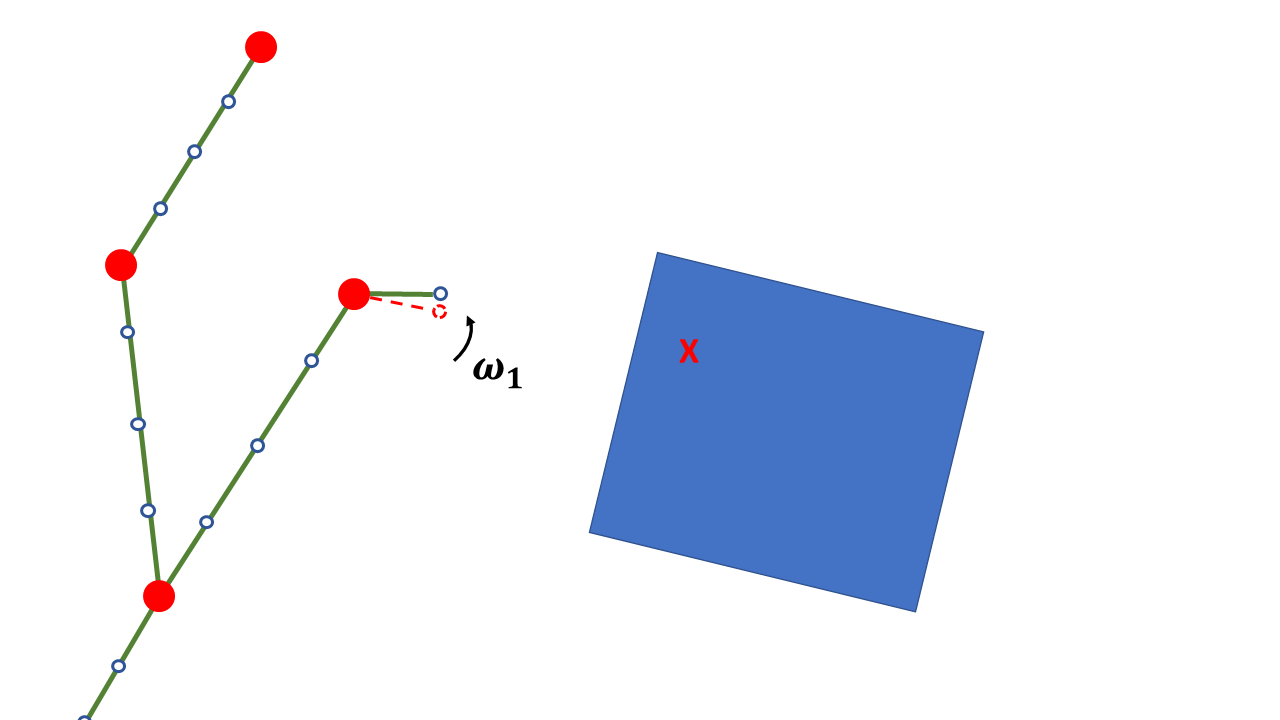}
         \caption{}
         \label{fig:step1}
     \end{subfigure}
     \begin{subfigure}[b]{0.32\linewidth}
         \centering
        \includegraphics[trim={0cm 0cm 0cm 0cm},clip,width=\linewidth]{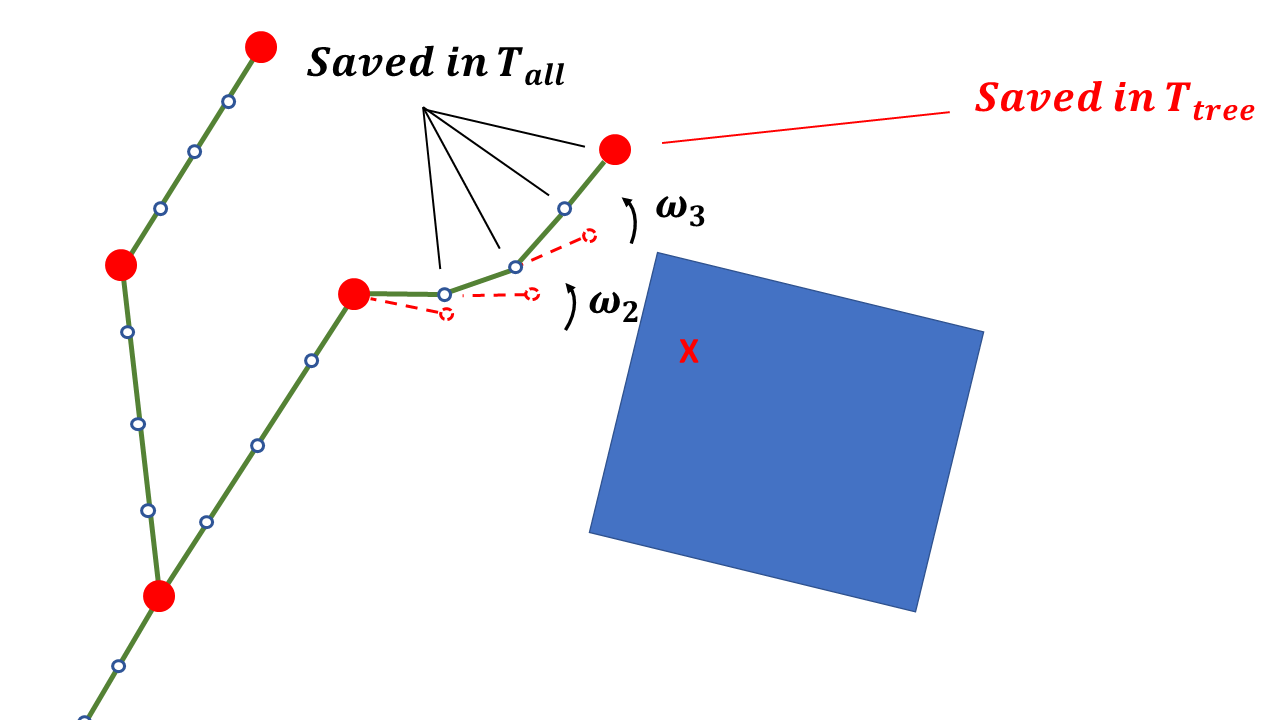}
         \caption{}
         \label{fig:step2}
     \end{subfigure}
\caption{\small{The green lines represent the existing path or tree. $T_{tree}$ only saves red nodes, and $T_{all}$ saves all nodes in the tree. The blue square is an obstacle. (a). Sample a point in a map and find the nearest node in $T_{tree}$ in the tree. (b). The red dashed point is the position that the robot plans to go. By solving CBF-QP in this position, a control input $\omega$ is generated and drives the robot in a new direction. (c). Repeat the process until all four steps have grown. The red node will be saved in $T_{tree}$, and four new nodes will be saved in $T_{all}$.}
} 
\vspace{-3mm}
% \label{fig:barrier exp}
\end{figure*}

\subsection{CBF-QP Safe Steering Controller}
% The robotic object of our study is a bipedal robot. 

Given a barrier function $h(\mathbf{x}) = \boldsymbol{\beta} \mathbf{z}^T$ generated by solving \eqref{eq:gradient_decend}, where $\mathbf{z}$ is composed of polynomial combinations of the states ($x_1, x_2$), we will formulate a CBF-QP steering control for the system defined in \eqref{eq:system_dynamics}.
Since $h(\mathbf{x})$ has relative degree two, we can construct a CBF according to (\ref{eq:cbf-cond}), given as
\begin{align}
    L_f^2 h(\mathbf{x}) + L_gL_fh(\mathbf{x})u+\alpha_2(\psi_1 (\mathbf{x})) \geq 0,
\end{align}
where
\begin{align}
    \psi_0 (\mathbf{x}) &= h(\mathbf{x}),\\
    \psi_1 (\mathbf{x}) &= \dot{\psi}_0 (\mathbf{x})+\alpha_1 (\psi_0 (\mathbf{x})).     
\end{align}
Following the definition of $h(\mathbf{x})$, we have
\begin{equation}
\begin{aligned}
    \dot{\psi}_0 (\mathbf{x}) = \dot{h}(\mathbf{x}) &= \boldsymbol{\beta}(\frac{\partial\mathbf{z}^T}{\partial x_1}\dot{x}_1+\frac{\partial\mathbf{z}^T}{\partial x_2}\dot{x}_2)\\
    &=\boldsymbol{\beta}(\frac{\partial\mathbf{z}^T}{\partial x_1}v\cos{\theta}+\frac{\partial\mathbf{z}^T}{\partial x_2}v\sin{\theta}),
\end{aligned}
\end{equation}
\begin{multline}
    L_f^2 h(\mathbf{x}) = \boldsymbol{\beta}(\frac{\partial^2\mathbf{z}^T}{\partial x_1^2}v^2\cos^2{\theta}+
    \frac{\partial}{\partial x_1}(\frac{\partial\mathbf{z}^T}{\partial x_2})v^2\sin{\theta}\cos{\theta}\\+\frac{\partial}{\partial x_2}(\frac{\partial\mathbf{z}^T}{\partial x_1})v^2\cos{\theta}\sin{\theta}+
    \frac{\partial^2\mathbf{z}^T}{\partial x_2^2}v^2\sin^2{\theta}),
\end{multline}
\begin{align}
    L_gL_fh(\mathbf{x})u =\boldsymbol{\beta}(-\frac{\partial\mathbf{z}^T}{\partial x_1}v\sin{\theta}+\frac{\partial\mathbf{z}^T}{\partial x_2}v\cos{\theta})u.
\end{align}
Hence, the resultant CBF constraint for a single obstacle $i$ is determined by:
\begin{multline}
% \begin{aligned}
    \varsigma_i(\mathbf{x},u) = \boldsymbol{\beta}_i(\frac{\partial^2\mathbf{z}^T}{\partial x_1^2}v^2\cos^2{\theta}+
    \frac{\partial}{\partial x_1}(\frac{\partial\mathbf{z}^T}{\partial x_2})v^2\sin{\theta}\cos{\theta}\\+\frac{\partial}{\partial x_2}(\frac{\partial\mathbf{z}^T}{\partial x_1})v^2\cos{\theta}\sin{\theta}+
    \frac{\partial^2\mathbf{z}^T}{\partial x_2^2}v^2\sin^2{\theta}\\
    +(-\frac{\partial\mathbf{z}^T}{\partial x_1}v\sin{\theta}+\frac{\partial\mathbf{z}^T}{\partial x_2}v\cos{\theta})u\\
    +k_1(\frac{\partial\mathbf{z}^T}{\partial x_1}v\cos{\theta}+\frac{\partial\mathbf{z}^T}{\partial x_2}v\sin{\theta})+k_0\mathbf{z}^T) \geq 0
% \end{aligned}
\end{multline}
where $k_0,k_1$ are user-determined CBF coefficients in the form of $k_1=\alpha_2,k_0=\alpha_0 \alpha_1$. If there are $N$ obstacles, the CBF-QP controller can be derived:
\begin{align}
\label{eq:cbf-qp-final}
    % \displaystyle 
    \min_{u} \hspace{1em} & || u-u_{ref} ||^2 \\
    \st & \varsigma_i(\mathbf{x},u) \geq 0 \quad \text{for } i=1,...,N_{obs} \nonumber\\
    & u_{\min} \leq u \leq u_{\max} \nonumber
\end{align}
where $u_{ref}$ is a reference angular velocity command.

\SetKwFunction{CP}{ChooseParent}
\SetKwFunction{Rw}{Rewrite}
\SetKwFunction{Sp}{Sampling}
\SetKwFunction{Near}{Nearest}
\SetKwFunction{At}{Atan2}
\SetKwFunction{CBFS}{CBFSteer}
\SetKwFunction{AD}{AddNode}
\SetKwFunction{FindN}{NearIndex}
\SetKwFunction{NG}{NearGoal}
\SetKwFunction{GP}{FinalPath}
\SetKwInput{Input}{Input}
\SetKwInput{Init}{Initialization}
\RestyleAlgo{ruled}

\SetKwFunction{Ex}{Extend}
\SetKwFunction{CBFQP}{CBF-QP}
\SetKwFunction{Sp}{Sampling}
\SetKwFunction{AU}{AngleUpdate}
\SetKwInput{Rt}{Return}

\RestyleAlgo{ruled}

\begin{algorithm}[t]
\vspace{3mm}
\caption{CBFSteer($\boldsymbol{\beta}, n_{near}, v, \theta_{x_s}, v$)}\label{alg:cbfsteer}
    \SetAlgoLined
    \Init{i=0, steps=4, $\theta = \theta_{x_s}$}
    \While{$i < steps$}{
        $i+=1$\;
        $n_{new} \xleftarrow[]{}$  \Ex($n_{near}, \theta, v$)\;
        $\omega \xleftarrow[]{}$  \CBFQP($n_{new}.x_1, n_{new}.x_2, \theta, v$)\;
        $\theta_{new} \xleftarrow[]{}$  \AU($\omega, \theta$)\;
        $n_{new} \xleftarrow[]{}$  \Ex($n_{near}, \theta_{new}, v$)\;
        $node\_list \xleftarrow{}$  \AD ($n_{new}$)\;
        $\theta = \theta_{new}$
    }
    \Rt{$n_{new}, node\_list$}
\end{algorithm}

\subsection{Implementation of the CBF-RRT* Algorithm}
Based on the CBF-QP, we formulate a steering procedure in RRT* every time it samples a new point on the map. 
Different from the steering method in the CBF-RRT* algorithm presented in \cite{https://doi.org/10.48550/arxiv.2206.00795}, our steering algorithm splits one big step into four small steps, and each of their headings is controlled by the angular velocity calculated through solving the CBF-QP in \eqref{eq:cbf-qp-final}. 
The structure of the steering method is shown in Algorithm \ref{alg:cbfsteer}.
Our proposed multi-step steering algorithm requires two node lists to save path data: tree node list $(T_{tree})$, that only saves the node after the tree has grown one big step in each loop; all node list $(T_{all})$, that saves all four nodes where the tree grows one big step, is used for tracking real motion trajectory. First, we sample points in the map, and select nearest node in $T_{tree}$, see \figref{fig:step0}. Then we calculate the position where the robot may go in the direction of the sample point for the first small step and solve its CBF-QP, which generates a control $\omega$ and may change the robot’s heading and walk to a new node, see \figref{fig:step1}. After that, we solve QP with the position along the new heading and repeat the same process until the tree has grown all four steps, see \figref{fig:step2}. All four newly generated nodes would be saved in $T_{all}$, and the last of four nodes would be saved in $T_{tree}$. By introducing this method, the proposed algorithm can effectively avoid the situation that CBFs become infeasible, and remain the robot safe.

% In order to get the relative optimal path, we need to integrate the CBF-QP steering method with RRT* algorithm. 
The overall procedure of the proposed CBF-RRT* is shown in Algorithm \ref{alg:cbf-rrt*}.
After saving the nodes into $T_{all}$ and $T_{tree}$, \CP and \Rw methods are used to change the nodes' parents to optimize the path. In particular, we use the simple collision-check function in these two procedures to improve the computational efficiency of the algorithm.
% This is because in determining whether the line between the two points touches the obstacle, using this function is equivalent to solving the CBF-QP which requires more computation. 
\begin{algorithm}
\caption{CBF-RRT*}\label{alg:cbf-rrt*}
    \SetAlgoLined
    \Input{$\mathcal{M}, \boldsymbol{\beta}, n_{init}, n_{goal}, N$} \algorithmiccomment{The map features, parameters of a barrier function, initial position, goal position, max iterations}\\
    \Init{i=0, $T_{all} = \{ n_{init} \}$, $T_{tree} = \{n_{init}\}, v = velocity$}
    \While{$i < N$}{
        $i+=1$\;
        $x_s \xleftarrow[]{}$  \Sp($\mathcal{M}$)\;
        $n_{near} \xleftarrow[]{}$  \Near($T_{tree}, x_s$)\;
        $\theta_{x_s} \xleftarrow[]{}$  \At($x_s, n_{near}$)\;
        $n_{new}, node\_list \xleftarrow{}$ \CBFS($\boldsymbol{\beta}, n_{near}, v, \theta_{x_s}, v$)\;
        $T_{tree} \xleftarrow{}$  \AD ($n_{new}$)\;
        \For{$node\_list$}{
            $Ind_{near} \xleftarrow{}$  \FindN ($each, T_{all}$)\;
            $each \xleftarrow{}$  \CP ($T_{all}, each, Ind_{near}$)\;
            $T_{all} \xleftarrow{}$  \AD ($each$)\;
            $T_{all} \xleftarrow{}$  \Rw ($T_{all}, each, Ind_{near}$)\;
            \uIf{\NG($each$)}{
                $path \xleftarrow{}$  \GP($T_{all}, each, n_{goal}$)\;
            }
        }
    }
\end{algorithm}

\section{Simulations and tests}
\label{sec:results}
% To demonstrate the effectiveness of our proposed work, we design several simulations and experimental tests with the ideal environment and real environment. 
We evaluate the effectiveness of the proposed work through several simulation and experimental tests. 
To start with, we simulated generating barrier functions and a safe path in an imagined environment with simple polygon shape obstacles. Then, we construct a map of an actual 
lab room using depth cameras and LiDAR of the Digit robot. The proposed algorithm can effectively identify obstacles in the middle of the room and surrounding desks and construct appropriate barrier functions. The CBF-RRT* then generates a safe path that enables the Digit robot to navigate between two chairs randomly placed in the room\footnote{Experiment recordings are shown in the following video: \url{https://youtu.be/r_hkuK5cMw4}}.

% After doing the simulation on a computer, we use a bipedal robot to do the path following phase to verify its feasibility. \par

% The robot we used for testing is called Digit, which is the product of AR company~\cite{castillo2021robust}. 

\subsection{Simulation Results}
\begin{figure}
\centering
% \vspace{2mm}
     \begin{subfigure}[b]{0.48\linewidth}
        %  \centering
        \includegraphics[trim={4cm 0cm 5cm 0cm},clip,width=\linewidth]{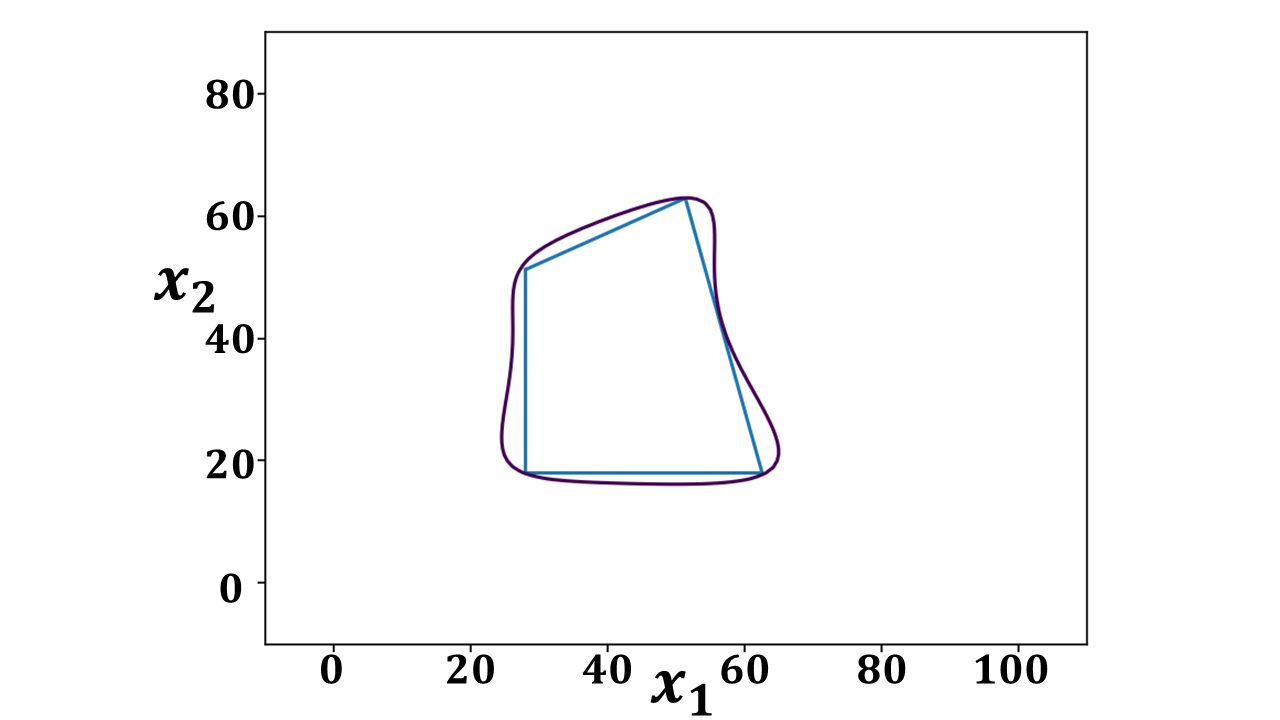}
         \caption{single obstacle}
         \label{fig:id1}
     \end{subfigure}
     \begin{subfigure}[b]{0.48\linewidth}
        %  \centering
        \includegraphics[trim={4cm 0cm 5cm 0cm},clip,width=\linewidth]{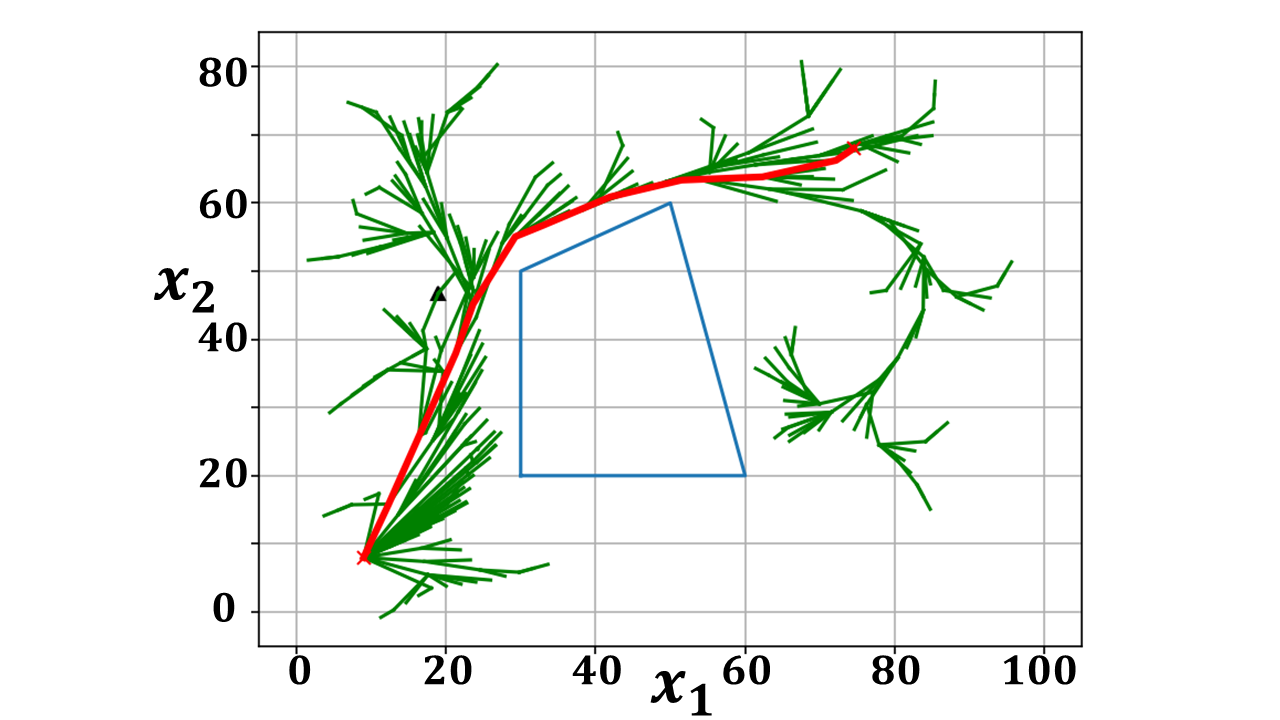}
         \caption{RRT tree and final path}
         \label{fig:id2}
     \end{subfigure}
     \begin{subfigure}[b]{0.48\linewidth}
        %  \centering
        \includegraphics[trim={3.8cm 0cm 4.5cm 0cm},clip,width=\linewidth]{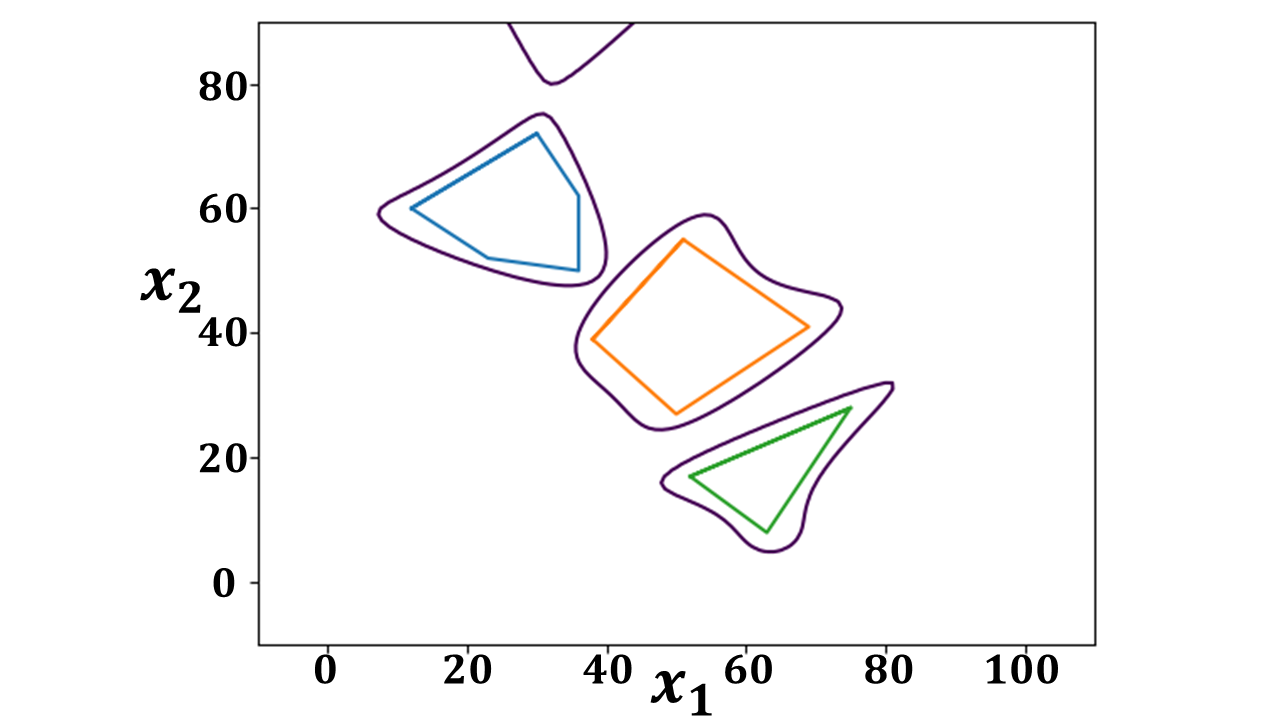}
         \caption{multiple obstacles}
         \label{fig:id3}
     \end{subfigure}
     \begin{subfigure}[b]{0.48\linewidth}
        %  \centering
        \includegraphics[trim={3.8cm 0cm 5cm 0cm},clip,width=\linewidth]{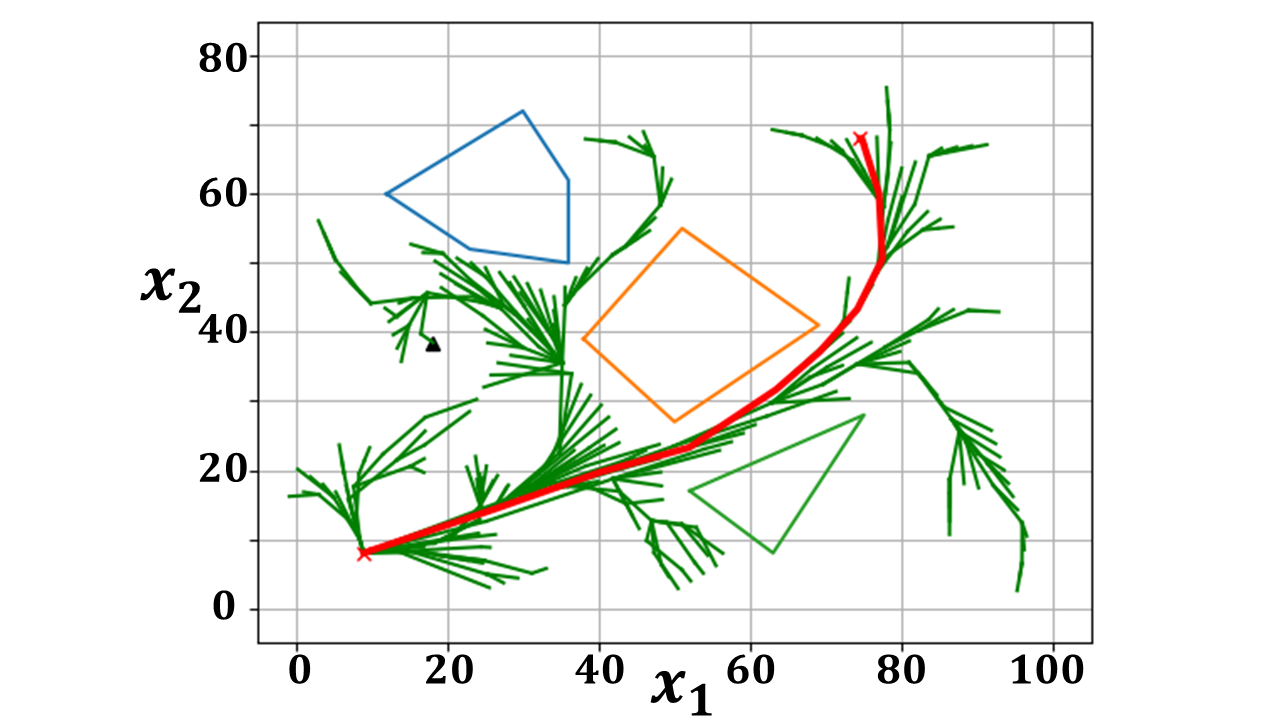}
         \caption{RRT tree and final path}
         \label{fig:id4}
     \end{subfigure}
\caption{\small{Simulated tests for polygon shape obstacles.}
} 
\label{fig:id testing}
\end{figure}

In this test, we consider two imaginary environments with $N_{obs} = 1$ and $N_{obs} = 3$, respectively. The obstacles and their barrier functions are shown in \figref{fig:id1} and \figref{fig:id3}. The time to construct barrier functions in each case is 0.64s and 1.84s. In order to keep the robot safe, we expand the obstacles with safety distance ($ds$). The rest of parameters and planning run time are shown in Table \ref{table:1}, including initial, goal states, CBF coefficients $k_0,k_1$. The path traced by CBF-RRT* is shown in \figref{fig:id2}, and \figref{fig:id4}. For the single-obstacle map, the relatively optimal path can be generated in 9.7s with 34 iterations. For the three-obstacle map, the relatively optimal path can be generated in 11.2s with 38 iterations.

\begin{table}[]
\centering
\begin{tabular}{||c c c c||}
\hline
\multicolumn{1}{|l|}{$k_0$} & \multicolumn{1}{l|}{$k_1$}  & \multicolumn{1}{l|}{$x_{init}$} & \multicolumn{1}{l|}{$x_{goal}$}   \\ \hline
\multicolumn{1}{|l|}{4} & \multicolumn{1}{l|}{2} &  \multicolumn{1}{l|}{(0.9, 0.8)m} & \multicolumn{1}{l|}{(7.45, 6.8)m}  \\ \hline\hline
\multicolumn{1}{|l|}{$v$} & \multicolumn{1}{l|}{$\omega_{ref}$} & \multicolumn{1}{l|}{$ds$} & \multicolumn{1}{l|}{$t_{planning}$} \\ \hline
\multicolumn{1}{|l|}{0.2 m/s} & \multicolumn{1}{l|}{0.0} & \multicolumn{1}{l|}{0.2 m}  & \multicolumn{1}{l|}{36.63s (N=1); 29.8s (N=3)} \\ \hline
\end{tabular}
\caption{The parameters and planning run times for ideal simulations.}
\label{table:1}
\end{table}

\subsection{Hardware Experiments with Digit}

Digit has an integrated perception system that includes three depth modules (Intel RealSense D430), one RGB-Depth module (Intel RealSense D435), one color camera (The Imaging Source DFM 27UP), and one LiDAR sensor (Velodyne LiDAR Puck VLP-16). The robot can walk robustly with a maximum velocity of 0.5 $m/s$. The locomotion can be controlled by either way-point commands or velocity commands. In our experiment, we used two pelvis depth modules and the LiDAR sensor to build the room map and used the way-point command to control the robot to walk.
% The robot’s torso contains Digit’s onboard computer, power source, and vision sensors. Vision sensors include three depth modules (Intel RealSense D430), one RGB-Depth module (Intel RealSense D435), one color camera (The Imaging Source DFM 27UP), and one LiDAR sensor (Velodyne LiDAR Puck VLP-16). The maximum velocity of the robot is 0.5 $m/s$, and its control inputs accept way-point command and velocity command. 

\begin{figure}
\centering
\vspace{2mm}
    \includegraphics[trim={4cm 0cm 4.5cm 0cm},clip,width=0.7\linewidth]{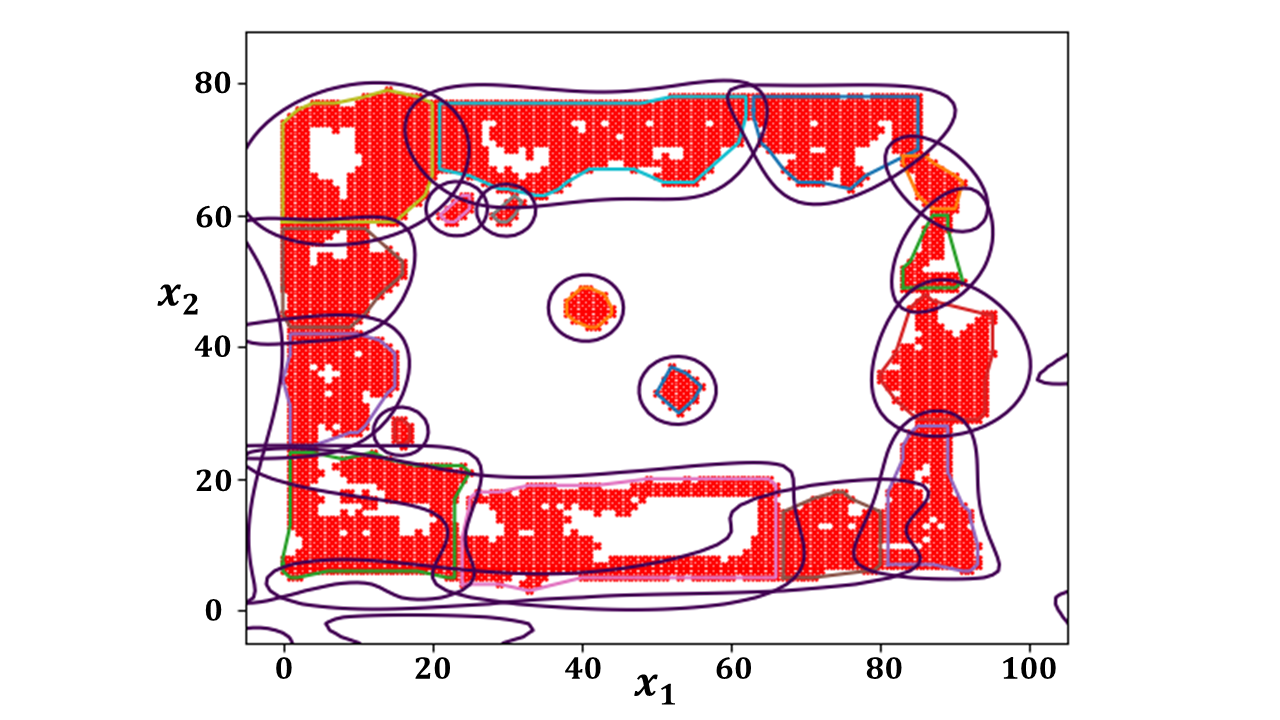}
     \caption{The generation of barrier functions from real-world 2D occupancy map.}
     \label{fig:rd1}
\end{figure}

\begin{figure}
         \centering
        \includegraphics[trim={4cm 0cm 4.5cm 0cm},clip,width=0.7\linewidth]{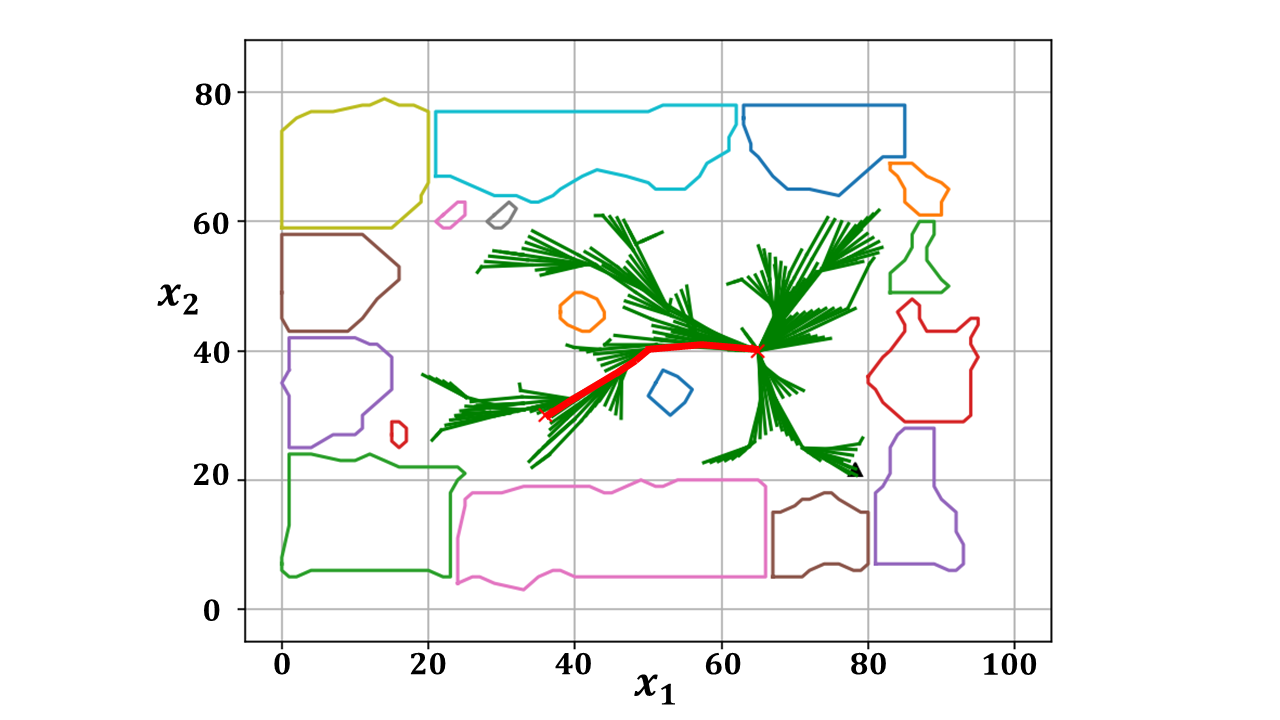}
         \caption{Collision-free safe path generated by the proposed CBF-RRT*.}
         \label{fig:rd2}
         \vspace{-2mm}
\end{figure}

In this test, we first need to generate an occupancy map of the room from the point cloud data obtained from the depth cameras and LiDAR. We used the Random Sample Consensus (RANSAC) method to segment point clouds to distinguish between obstacles and free space. The resulting obstacle points are projected onto a 2D plane, and the cells containing projected points are considered occupied by obstacles. To improve the efficiency of constructing barrier functions, we divided the map into multiple regions and generated the barrier functions for each region respectively. \figref{fig:rd1} shows the occupancy map of the room and generated barrier functions of obstacles. The time to construct these obstacles is 3.03s. We set the maximum simulation iterations to 120. Given the initial and goal position, our proposed CBF-RRT* generated a safe path passing through two obstacles in the middle of the room in 12.54s with 37 iterations, as shown in \figref{fig:rd2}. The parameters and planning run time of the real test is shown in Table \ref{table:2}. In the path following phase, we set Digit's velocity to $0.1 m/s$, and send the way-point in the path to Digit. \figref{fig:rd3} shows the snapshots of the robot following the path and avoiding obstacles. 
% The robot uses 48s to finish this path.

\begin{table}[]
\centering
\begin{tabular}{||c c c c||}
\hline
\multicolumn{1}{|l|}{$k_0$} & \multicolumn{1}{l|}{$k_1$}  & \multicolumn{1}{l|}{$x_{init}$} & \multicolumn{1}{l|}{$x_{goal}$}   \\ \hline
\multicolumn{1}{|l|}{4} & \multicolumn{1}{l|}{2} &  \multicolumn{1}{l|}{(6.5, 4.0)m} & \multicolumn{1}{l|}{(3.6, 3.0)m}  \\ \hline\hline
\multicolumn{1}{|l|}{$v$} & \multicolumn{1}{l|}{$\omega_{ref}$} & \multicolumn{1}{l|}{$ds$} & \multicolumn{1}{l|}{$t_{planning}$} \\ \hline
\multicolumn{1}{|l|}{0.1 m/s} & \multicolumn{1}{l|}{0.0} & \multicolumn{1}{l|}{0.15 m}  & \multicolumn{1}{l|}{37.94s} \\ \hline
\end{tabular}
\caption{The parameters and planning running time for real test.}
\label{table:2}
\vspace{-3mm}
\end{table}


\subsection{Analysis}
One of the major focuses of this work is avoiding the polynomial shape obstacles and generating a safety path on CBF-based RRT*. Through the above simulations and experimental tests, we validated that the proposed algorithm can construct appropriate CBFs for arbitrary shape obstacles and generate a collision-free path effectively. Also, this algorithm can find a relatively optimal path within 15 seconds and 40 interactions. Compared to other CBF-RRT/RRT* methods, our method can consider efficiency while making the path relatively more optimized. In the real environment test, it is shown that the resulting path is feasible for Digit. However, we also notice that Digit's walking velocity is relatively slow in the path following phase, and the following progress of Digit is not very coherent, which may be due to the way-point command.

% , or it may also be caused by the simplified dynamic model. 

\vspace{-3mm}
\section{Conclusions}
\label{sec:conclusion}
In this paper, we introduced a new framework of CBF-RRT* with the ability to generate a collision-free path for complex-shaped obstacles. We also demonstrated the feasibility of the algorithm in real-world environments through hardware experiments. 
% Although our algorithm is feasible for many polygon shape obstacles, there are still difficulties in fitting more complex shapes, such as hollow-shaped obstacles. 
In future work, we plan to construct CBFs for more complex shape obstacles to improve the better performance in practice. We will further improve the computational efficiency of the algorithm so that it can also be applied to safe navigation in dynamic environments.
% Also, in the simulation, our obstacles are relatively static, so in future research, we hope to explore the CBFs of moving polygon shape obstacles and safety path planning as well.
% Moreover, during the planning process, we used a simplified dynamics model of the biped robot. Although it can basically meet the control requirements, we still hope to use a sophisticated model to achieve more accurate results. Thus, future research will also focus on the CBF-RRT* integration with the dynamics of bipedal robots.
Moreover, the differential drive model may limit the agility of bipedal robots. Hence, we will explore more suitable model representations for path planning that unlocks the potential of bipedal walking robots. 

\newpage
\bibliographystyle{IEEEtran}
\bibliography{references.bib}

\end{document}